\documentclass[acmtog]{acmart}

\usepackage{booktabs} 

\usepackage[ruled]{algorithm2e} 

\SetAlFnt{\small}
\SetAlCapFnt{\small}
\SetAlCapNameFnt{\small}
\SetAlCapHSkip{0pt}

\usepackage{xcolor}

\acmJournal{TOG}

\acmDOI{3757377.3763812}

\copyrightyear{2025}
\acmYear{2025}
\setcopyright{acmlicensed}\acmConference[SA Conference Papers '25]{SIGGRAPH Asia 2025 Conference Papers}{December 15--18, 2025}{Hong Kong, Hong Kong}
\acmBooktitle{SIGGRAPH Asia 2025 Conference Papers (SA Conference Papers '25), December 15--18, 2025, Hong Kong, Hong Kong}
\acmDOI{10.1145/3757377.3763812}
\acmISBN{979-8-4007-2137-3/2025/12}

\newcommand*{\MethodName}{ShapeGen}

\begin{document}
\title{\MethodName{}: Towards High-Quality 3D Shape Synthesis}

\author{Yangguang Li}
\authornote{Equal contribution.}
\orcid{0000-0002-6090-3899}
\affiliation{%
 \institution{The Chinese University of Hong Kong, VAST}
 \country{Hong Kong}}
\email{liyangguang256@gmail.com}

\author{Xianglong He}
\authornotemark[1]
\orcid{0000-0001-5368-0765}
\affiliation{%
 \institution{Tsinghua University, VAST}
 \country{China}}
\email{hxl23@mails.tsinghua.edu.cn}

\author{Zi-Xin Zou}
\orcid{0000-0003-2945-552X}
\affiliation{%
 \institution{VAST}
 \country{China}}
\email{zouzx1997@gmail.com}

\author{Zexiang Liu}
\orcid{0009-0007-8287-8499}
\affiliation{%
 \institution{VAST}
 \country{China}}
\email{liuzexiang1201@gmail.com}

\author{Wanli Ouyang}
\orcid{0000-0002-9163-2761}
\affiliation{%
 \institution{The Chinese University of Hong Kong, Shanghai Artificial Intelligence Laboratory}
 \country{Hong Kong}}
\email{wlouyang@ie.cuhk.edu.hk}

\author{Ding Liang}
\orcid{0000-0001-9774-4687}
\authornote{Corresponding author.}
\affiliation{%
 \institution{VAST}
 \country{China}}
\email{liangding1990@163.com}

\author{Yan-Pei Cao}
\orcid{0000-0002-0416-4374}
\authornotemark[2]
\affiliation{%
 \institution{VAST}
 \country{China}}
\email{caoyanpei@gmail.com}


%
%
\ccsdesc[500]{Computing methodologies~Computer graphics}
\ccsdesc[500]{Computing methodologies~Shape modeling}
\ccsdesc[500]{Computing methodologies~Mesh geometry models
}

%
%

\begin{teaserfigure}
    \centering
    \vspace{-1em}
    \includegraphics[width=\textwidth]{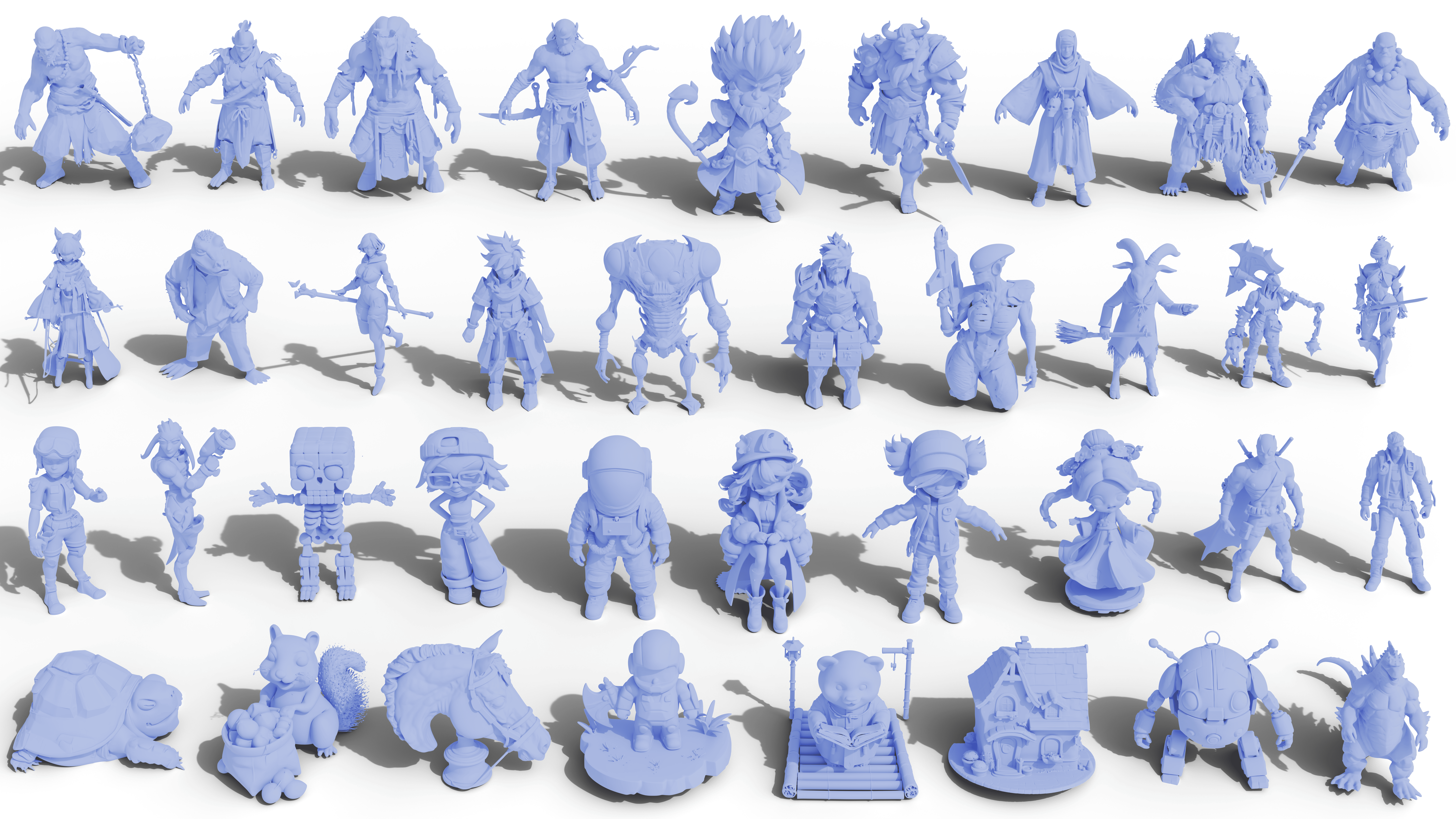}
    \vspace{-2em}
    \caption{Overall results of \MethodName{}. Our method demonstrates strong generalization and generates high-quality, detailed 3D assets.
    }
\label{fig:teaser}
\end{teaserfigure}

\begin{abstract}
Inspired by generative paradigms in image and video, 3D shape generation has made notable progress, enabling the rapid synthesis of high-fidelity 3D assets from a single image. 
However, current methods still face challenges, including the lack of intricate details, overly smoothed surfaces, and fragmented thin-shell structures. These limitations leave the generated 3D assets still one step short of meeting the standards favored by artists.
In this paper, we present \MethodName{}, which achieves high-quality image-to-3D shape generation through 3D representation and supervision improvements, resolution scaling up, and the advantages of linear transformers. These advancements allow the generated assets to be seamlessly integrated into 3D pipelines, facilitating their widespread adoption across various applications.
Specifically, in contrast to existing methods: 
1) We investigate how different representations and VAE supervision strategies affect the generation process, and address issues like aliasing artifacts and fragmented thin-shell structures by using an TSDF-based representation supervised with BCE loss.
2) We scale up the resolution of 3D data, image conditioning inputs, and the number of latent tokens to enhance generation fidelity. 
3) We adopt mixed conditioning using raw RGB images and normal maps during training, effectively resolving ambiguities caused by inconsistencies between ControlNet-generated RGB images and the underlying geometry from untextured assets.
4) We replace the original softmax attention with linear attention to improve training and inference efficiency when handling a large number of latent tokens.
5) We introduce an inference-time scaling strategy that enhances generation quality at test time.
Through extensive experiments, we validate the impact of these improvements on overall performance. Ultimately, thanks to the synergistic effects of these enhancements, \MethodName{} achieves a significant leap in image-to-3D generation, establishing a new state-of-the-art performance.

\end{abstract}

\keywords{Image-to-3D Generation, Shape Generation, Shape Modeling}

\maketitle
\section{Introduction}
Inspired by recent advances in image and video generation architectures and training paradigms, 3D asset generation has witnessed substantial advancements. Representative 3D generative foundation models, such as ~\cite{zhang2024clay,zhao2025hunyuan3d,li2025triposg,chen2024dora}, build upon the framework introduced by 3DShape2VecSet~\cite{zhang20233dshape2vecset}, which first compresses raw 3D representations into a latent space via a 3D Variational Autoencoder (3D-VAE), and then leverages a Diffusion Transformer (DiT) architecture~\cite{peebles2023scalable}, coupled with advanced sampling strategies, to train image-conditioned 3D generation models.
Unlike previous methods~\cite{long2024wonder3d,liu2023syncdreamer} that only generate suboptimal 3D assets using multi-view images as intermediate representations, these pre-trained foundation models~\cite{zhang2024clay,zhao2025hunyuan3d,li2025triposg} enable the rapid synthesis of high-fidelity 3D assets that are semantically aligned with the content of a single input image from arbitrary viewpoint. This capability marks a paradigm shift in the 3D content creation pipeline, offering a scalable and efficient alternative to traditional manual modeling.

Despite the remarkable capabilities of current 3D foundation models that have captivated the 3D modeling community, they continue to face challenges due to suboptimal choices in data processing, architectural design, and training strategies. For example, the suboptimal representation introduced during VAE training often results in fragmented thin-shell structures~\cite{li2025triposg}, aliasing or “staircasing” artifacts in DiT generation~\cite{zhang2024clay,chen2024dora}. Additionally, limitations such as low-resolution watertight data, low-resolution conditioning images, and a limited number of latent tokens contribute to a lack of fine-grained details in the generated 3D assets.
These constraints collectively hinder the visual fidelity and structural coherence of the outputs, leaving a noticeable gap between the generated models and the quality standards typically demanded by professional artists. 

In this paper, we present \MethodName{}, a novel framework for 3D generation that systematically analyzes the limitations of existing approaches and introduces targeted improvements to address them. Our method not only overcomes the shortcomings of prior models but also enables the generation of higher-fidelity and higher-quality 3D assets.
The key improvements of \MethodName{} are listed as follows:

\textbf{VAE Representation and Supervision.} 
We analyze how different VAE representations and supervision schemes affect flow-based 3D generation. 
While occupancy representations trained with BCE supervision improve flow models training stability but introduce aliasing or “staircasing” artifacts, Signed Distance Function (SDF) representations with MSE supervision yield smoother surfaces but fragmented thin-shell structures. 
To overcome these limitations, following Surf-D~\cite{yu2024surf}, we incorporate a BCE loss on the truncated SDF-based representation to enhance robustness against latent noise.
It combines the advantages of both occupancy and SDF representation, enabling flow models to generate 3D shapes with smooth surfaces and structurally coherent thin shells.

\textbf{Resolution Improvement.} 
We improve generation quality by scaling up system resolution in terms of 3D data processed, image condition, and latent token number. Specifically, we use $1024^3$ 3D data resolution, $518\times518$ image resolution, and $65k$ tokens to enhance geometry fidelity, capture finer image semantics, and improve reconstruction and generation capacity. 

\textbf{Mixed Condition Training.} 
To address inconsistencies in training data caused by missing textures or synthesizing textures by ControlNet~\cite{zhang2023adding} in Objaverse(-XL)~\cite{Deitke_2023_CVPR,deitke2023objaverse} datasets, we use a mixed condition training strategy. Specifically, we use RGB images for textured assets and normal maps for untextured ones, ensuring geometry-aligned and appearance-consistent supervision. This reduces ambiguity during training and leads to higher-fidelity 3D generation.

\textbf{Linear Attention.} 
While prior works like SANA~\cite{xie2024sana} adapt linear attention for high-resolution image generation, they face locality degradation due to local spatial correlations in compressed image tokens. SANA replaces the original FFN with a convolutional MixFFN to recover locality, but it is at the cost of offsetting the speed gains brought by linear attention.
In contrast, 3D tokens derived from 1D sequences naturally encode global structure, allowing us to adopt linear attention based on the LightNet~\cite{qin2024you} design without sacrificing fidelity. This enables efficient scaling to large token number for faster 3D generation.

\textbf{Inference-Time Scaling.} 
Inspired by the success of inference-time scaling in LLMs and image generations~\cite{wang2022self,zhuo2025reflection}, we apply an inference-time scaling technique~\cite{ma2025inference} during the sampling phase of our generation model to improve the quality of generation. Specifically, we incorporate an auxiliary evaluation model during inference to assess the quality of multiple generated candidates, iteratively refining candidates with their geometry quality, searching and selecting the one associated with the highest-quality seed. The strategy does not require additional training and only introduces a modest increase in inference time, while yielding noticeable improvements in geometric fidelity.

Building upon the base VAE and generation architecture introduced in 3DShape2VecSet~\cite{zhang20233dshape2vecset} and TripoSG~\cite{li2025triposg}, we integrate all of the aforementioned improvements to reach \MethodName{}. 
With these improvements, \MethodName{} achieves a new state-of-the-art in the image-to-3D generation task, as shown in Fig.\ref{fig:teaser}, \ref{fig:shapegen_comparison} and \ref{fig:shapegen_demo_show}. It not only addresses the limitations, such as aliasing or “staircasing” artifacts and fragmented thin-shell structures, of existing approaches, but also establishes a new SOTA performance in terms of shape fidelity, image-shape consistency, and overall generation quality.
\section{Related Works}
\paragraph{3D Large-scale Dataset.} 
High-quality, large-scale 3D datasets are fundamental to the success of current 3D asset generation models, especially those based on data-hungry paradigms like diffusion. ShapeNet~\cite{chang2015shapenet}, one of most influential 3D datasets, offers around 51,300 models across 55 categories, along with rich semantic annotations. While widely adopted, its limited scale constrains its applicability to large generative models. Objaverse~\cite{Deitke_2023_CVPR} and its extension Objaverse-XL~\cite{deitke2023objaverse} significantly expand the data frontier, providing about 800K and 10M assets respectively, and support the exploration of scaling laws in 3D generation. However, Objaverse-XL presents challenges such as various formats, noisy or inaccurate annotations. Addressing these limitations through data cleaning, curation, and filtering remains crucial for unlocking the full potential of large-scale 3D generation.

\paragraph{3D Representation.}
A variety of 3D representations have been explored for generative model, each offering trade-offs between flexibility, fidelity, and computational costs. Point clouds are lightweight and easy to process, making them attractive for generation tasks. Early works like PointNet~\cite{qi2017pointnet} laid the foundation for neural processing of point sets, while recent diffusion-based models~\cite{luo2021diffusion,nichol2022point,vahdat2022lion} treat points as samples from learnable distributions. However, point clouds lack explicit connectivity and surface continuity, often requiring post surface reconstruction (e.g., Marching Cubes~\cite{lorensen1998marching}). Triangle meshes provide a rich, structured representation widely used in industry. Generative methods either predict intermediate fields and convert them to meshes via iso-surfacing~\cite{kerbl20233d,park2019deepsdf}, or directly autoregress mesh elements~\cite{nash2020polygen,siddiqui2024meshgpt,tang2024edgerunner}. Diffusion-based mesh models like MeshCraft~\cite{he2025meshcraft} further enhance geometry synthesis, though producing high-resolution, topologically sound meshes remains challenging. Volumetric fields, including occupancy and SDF-based representations~\cite{choy20163d,brock2016generative,wu2016learning}, offer dense spatial encoding but suffer from cubic scaling in memory with resolution. Octree-based schemes~\cite{wang2022dual} and hybrid voxel-SDF pipelines mitigate this, enabling coarse-to-fine refinement. Finally, neural fields represent shapes as continuous functions using coordinate-based networks~\cite{zhang20233dshape2vecset,zhao2023michelangelo,zhang2024clay}, allowing topology-agnostic modeling. These methods often require isosurface extraction and watertight conversion, and face scalability challenges due to memory demands at high resolution. Each representation contributes uniquely to 3D generation, influencing the choice of architecture and training strategy.

\paragraph{3D Generation Model.}
Early 3D generation used classic paradigms such as VAEs~\cite{kingma2013auto}, GANs~\cite{goodfellow2014generative}, and autoregressive models~\cite{gregor2014deep}, applied to voxels, point clouds, or meshes~\cite{wu2016learning, sanghi2022clip, yan2022shapeformer}. With the rise of 2D diffusion models like Imagen~\cite{saharia2022photorealistic} and Stable Diffusion~\cite{rombach2022high}, methods such as DreamFusion~\cite{poole2022dreamfusion} introduced score distillation sampling(SDS) to optimize NeRF without 3D supervision. Although extended to neural fields including DMTet~\cite{shen2021deep} and 3D Gaussian Splatting~\cite{kerbl20233d}, SDS-based methods remain slow and lack explicit 3D priors.
To overcome this, recent works incorporate 3D priors through either large-scale reconstruction model\cite{hong2023lrm, tochilkin2024triposr, zou2024triplane} or direct 3D generation. The former predicts NeRF from single or multiview inputs, but often yields coarse geometry. The latter trains diffusion or autoregressive models directly on 3D data. Examples include NeuralWavelet~\cite{hui2022neural} for frequency-domain diffusion, Point-E~\cite{nichol2022point} for two-stage point cloud synthesis, and MeshGPT~\cite{siddiqui2024meshgpt} and MeshCraft~\cite{he2025meshcraft} for mesh generation. Recent models like Hunyuan3D-2~\cite{zhao2025hunyuan3d} and TripoSG~\cite{li2025triposg} further scale up transformer-based architectures for image-conditioned 3D synthesis.
Despite these advances, challenges such as missing details, smooth surfaces, and disconnected structures persist. Our method addresses these issues via enhanced supervision, resolution scaling, and hybrid conditioning, achieving state-of-the-art results.
\section{Method}
\subsection{Preliminary}\label{sec:Preliminary}
The prior work 3DShape2VecSet~\cite{zhang20233dshape2vecset} introduced an effective framework for the representation and compression of 3D assets. Raw geometric assets are first converted into watertight models to ensure topological consistency. Subsequently, two types of point samples are constructed for the input and supervision of VAE model training: (1) it takes a uniform surface point cloud $X$ of size $N\times6$ as input, with each point comprising its spatial coordinate and an associated surface normal;
(2) supervision is provided by $M\times4$ points randomly sampled within the continuous volume space $R^3$ containing the model, where each point comprises its spatial coordinates and a label encoding geometry information. 
This design enables the construction of an implicit-based representation that faithfully captures the underlying 3D structure.

Based on this representation, the encoder architecture of the 3D VAE, as described in Equations~(1)--(2), proceeds as follows. Given the input \( X \), a downsampled set \( X' \) is first obtained using Farthest Point Sampling (FPS). Both \( X \) and \( X' \) are then embedded through $\text{PosEmb}$ and processed by a $\text{CrossAttn}$ module, where geometric information from \( X \) is learned to \( X' \), resulting in \( Z_0 \). Subsequently, \( Z_0 \) is encoded and compressed into a low-dimensional latent feature \( Z \) through $i$ layers  of $\text{SelfAttn}$ followed by a $\text{Linear}$ layer. 
\begin{equation}\label{encoder_equation}
\mathbf{Z}_0 = \text{CrossAttn}(\text{PosEmb}(\mathbf{X}), \text{PosEmb}(\mathbf{X'}))
\end{equation}
\begin{equation}
\mathbf{Z} = \text{Linear}\left( \text{SelfAttn}^{(i)}(\mathbf{Z}_0) \right)
\end{equation}

The decoder architecture of the 3D VAE, as described in Equations~(3)--(4). Specifically, the latent feature \( Z \) is first projected to a higher-dimensional space through a $\text{Linear}$ layer. It is then decoded into \( \tilde{Z} \) by applying \( j \) layers of $\text{SelfAttn}$. 
Query points \( x \) discretely sampled from the 3D space at different resolutions are processed by $\text{PosEmb}$ and attend to \( \tilde{Z} \) via a $\text{CrossAttn}$ module to predict occupancy values across the field.
Finally, the 3D geometry will be extracted from the occupancy field using the Marching Cubes.
\begin{equation}
\tilde{\mathbf{Z}} = \text{SelfAttn}^{(j)}\left( \text{Linear}(\mathbf{Z}) \right)
\end{equation}
\begin{equation}
s = \text{CrossAttn}\left( \text{PosEmb}(x), \tilde{\mathbf{Z}} \right)
\end{equation}

In the generative model, the DiT~\cite{peebles2023scalable} is employed to learn the mapping between the noise distribution and the latent feature \( Z \) under image conditioning. The goal is to train a generation model capable of sampling high-fidelity 3D geometry assets from noise guided by different input images as conditions. Besides, DINOv2~\cite{oquab2023dinov2} is utilized to extract image features that are injected as conditioning information. For noise sampling strategies, previous works have explored alternatives such as EDM~\cite{karras2022elucidating}, DDPM~\cite{ho2020denoising} and Rectified Flow~\cite{liu2022flow}.

\subsection{Overview}
Building upon the 3D VAE framework and Rectified Flow noise sampling strategy validated in prior works~\cite{zhang20233dshape2vecset,li2025triposg}, this paper further explores the effects of several key factors on overall generation performance. We begin by presenting the overall architecture of \MethodName{}, as detailed in Sec.\ref{sec:Architecture}, and subsequently investigate: (1) the differences in VAE representation and supervision strategies from a generative modeling perspective, as detailed in Sec.\ref{sec:VAE Supervision}; (2) resolution scaling up from different processes, as detailed in Sec.\ref{sec:Resolution Improvement}; (3) the use of mixed condition training with RGB and Normal maps, as detailed in Sec.\ref{sec:Hybrid Condition}; (4) the effects of linear attention mechanisms on 3D shape generation, as detailed in Sec.\ref{sec:Linear Attention}; and (5) inference-time scaling approaches for enhancing generation quality, as detailed in Sec.\ref{sec:Inference Scaling}.

\subsection{Architecture}\label{sec:Architecture}
\begin{figure}
    \centering
    \includegraphics[width=1.0\linewidth]{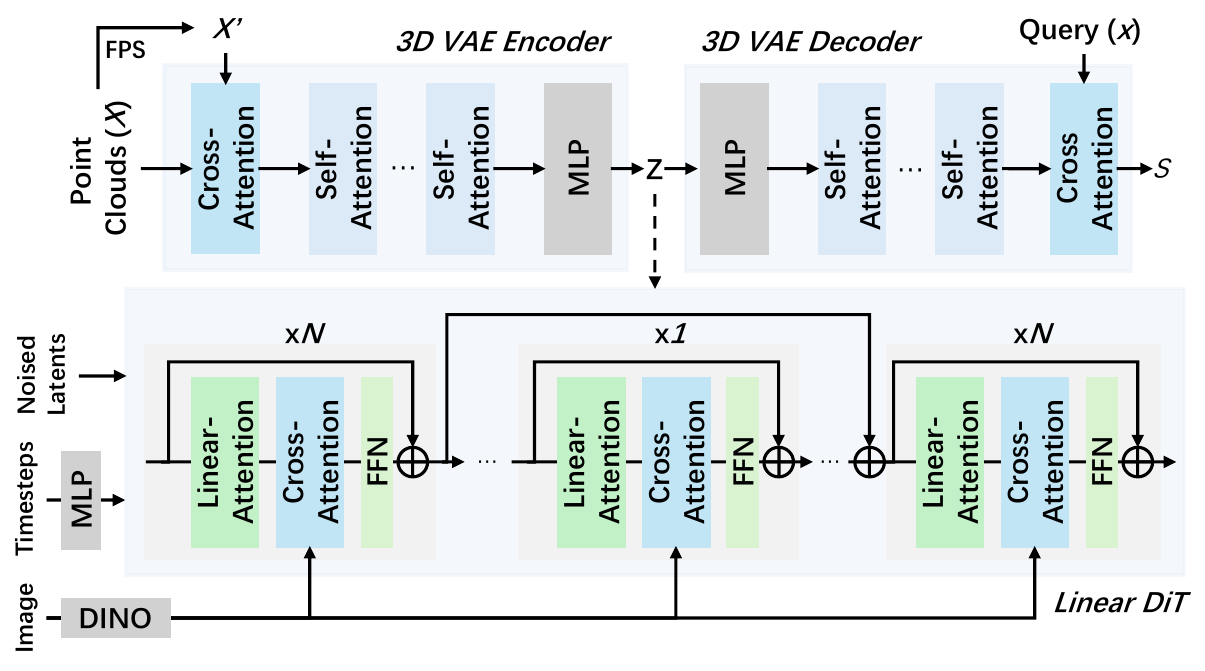}
    \vspace{-1em}
    \caption{The overall architecture of \MethodName{}.}
    \label{fig:overall-architecture}
\end{figure}
As shown in the Fig.\ref{fig:overall-architecture}, the 3D VAE architecture follows the structure introduced in the preliminary, as detailed in sec.\ref{sec:Preliminary}, without structural modifications. Specifically, the encoder employs \( i = 8 \) layers of $\text{SelfAttn}$ with a hidden dimension of $512$, while the decoder uses \( j = 16 \) layers of $\text{SelfAttn}$ with a hidden dimension of $1024$. The latent feature \( Z \) has a dimensionality of $64$. 
For the input, we use a surface point cloud consisting of \( X = 204{,}800 \) points, which is downsampled to a maximum token number of \( X' = 65{,}536 \) using FPS. For the output, we discretize the 3D space into a \( 512^3 \) grid and perform SDF field queries at each grid point to reconstruct the 3D geometry.

In the generation architecture, we adopt the linear attention module introduced in LightNet~\cite{qin2024you} as the basic DiT block, combined with the skip-connection mechanism proposed in U-ViT~\cite{bao2023all} and further validated in TripoSG~\cite{li2025triposg}, to construct the backbone of our model. The overall architecture consists of $21$ layers, including $10$ encoder blocks, $1$ middle block, and $10$ decoder blocks. 
For conditional input, we extract image features at a resolution of $518$ using DINOv2~\cite{oquab2023dinov2} register version, and inject them into the model via a $CrossAttn$ module. The time-step information is encoded through $MLP$ layers and concatenated with the input latent representation. Following the design of TripoSG, we employ Rectified Flow~\cite{liu2022flow} as the sampling strategy. 
Based on these components, we construct the overall generation architecture of \MethodName{}.

\subsection{VAE Representation and Supervision}\label{sec:VAE Supervision}
We systematically analyze the occupancy and SDF representation and supervision in VAEs adopted by existing works. Our flow-based experiments reveal that, although these VAEs exhibit similar reconstruction performance, they lead to clear differences in flow models training stability and generation quality. 
Specifically, VAEs trained with occupancy representation with BCE supervision facilitate better convergence during flow model training, yet the resulting 3D shapes often exhibit aliasing or “staircasing” artifacts, which may be due to the discrete nature of occupancy. In contrast, VAEs trained with continuous SDF representation with MSE supervision produce more visually appealing and smooth surfaces, but are more prone to generating fragmented or inconsistent thin-shell structures, which is an issue particularly noticeable in semantically complex or concave geometries.
To address these limitations, we convert MSE supervision on SDF representation to a BCE supervision inspired by Surf-D~\cite{yu2024surf} and enhance VAE training with additional normal signal supervision. 
The VAE predicts the SDF, while surface normals are derived through PyTorch's automatic differentiation framework. This dual-supervision strategy, which enforces constraints on both the scalar values of the SDF field and its spatial gradients, inherently guarantees geometric consistency between the reconstructed SDF and the corresponding normal vectors.
The computation process is illustrated in Equations~(\ref{eq:clip})--(\ref{eq:recon}). Specifically, Equation~(\ref{eq:clip}) clips the ground-truth SDF $y$ by a truncation threshold $\delta$, and transforms it from range $[-1,1]$ in the range $[0, 1]$ to obtain supervision label $\tilde{y}$. 
In Equation~(\ref{eq:sdf}), the predicted logit $\hat{x}$ is passed through a sigmoid function $\sigma(\cdot)$, and the BCE loss is computed between $\sigma(\hat{x})$ and $\tilde{y}$ to form  $\mathcal{L}_{\text{bce-sdf}}$. 
Equation~(\ref{eq:normal}) computes the MSE loss between the predicted surface normals $\hat{\mathbf{n}}$ and the ground-truth normals $\mathbf{n}$, resulting in $\mathcal{L}_{\text{normal}}$. 
Finally, Equation~(\ref{eq:recon}) combines these two components via a weighted sum to obtain the overall reconstruction loss $\mathcal{L}_{\text{recon}}$.
\begin{equation}\label{eq:clip}
\tilde{y} = \frac{\text{clip}(y, -\delta, \delta) + \delta}{2\delta}
\end{equation}
\vspace{-2em}

\begin{align}\label{eq:sdf}
\mathcal{L}_{\text{bce-sdf}} &= - \left[ \tilde{y} \cdot \log \sigma(\hat{x}) + (1 - \tilde{y}) \cdot \log(1 - \sigma(\hat{x})) \right]
\end{align}
\vspace{-2em}

\begin{equation}\label{eq:normal}
\mathcal{L}_{\text{normal}} = \left\| \hat{\mathbf{n}} - \mathbf{n} \right\|_2^2
\end{equation}
\vspace{-2em}

\begin{equation}\label{eq:recon}
\mathcal{L}_{\text{recon}} = \lambda_{\text{bce}} \cdot \mathcal{L}_{\text{bce-sdf}} + \lambda_{\text{normal}} \cdot \mathcal{L}_{\text{normal}}
\end{equation}
This supervision scheme effectively combines the advantages of both occupancy and SDF representation while mitigating their respective drawbacks.
Flow models trained with these improved VAEs are capable of generating 3D shapes with smooth surfaces and structurally coherent thin shells.

\subsection{Resolution Improvement}\label{sec:Resolution Improvement}
To improve the quality of 3D generation, we systematically scale up the resolution of three critical components in the pipeline: 3D data representation, image condition, and latent token number.

Previous works~\cite{zhang2024clay,li2025triposg} typically pre-process raw 3D assets into watertight occupancy or SDF representations at a resolution of $512^3$, which introduces quantization artifacts and blurs fine geometric details. In contrast, we adopt a higher resolution of $1024^3$ during data processing and model training. 
The resolution of the input image plays a crucial role in the generation process. While prior methods~\cite{zhang2024clay,li2025triposg} resize the conditioning image to $224\times224$, we increase the resolution to $518\times518$, enabling the model to extract and leverage richer semantic and structural cues from the input. 
To encode and decode high-resolution 3D representations, we scale the number of latent tokens in the VAE from the conventional 2K–4K~\cite{zhang2024clay,li2025triposg,zhao2025hunyuan3d} range up to 65K tokens. This substantial increase expands the representational capacity of the latent space, allowing the model to better preserve local and global structures during reconstruction and generation. 

Together, these scaling strategies form the foundation of our high-fidelity 3D generation pipeline, enabling significant performance gains over existing approaches.

\subsection{Mixed Condition Training}\label{sec:Hybrid Condition}
Existing 3D generative models are predominantly trained on the Objaverse(-XL)~\cite{Deitke_2023_CVPR,deitke2023objaverse} dataset, where a large portion of assets lack texture. While the data processing pipeline proposed in TripoSG~\cite{li2025triposg} allows the generation of multi-view RGB images for untextured assets to serve as image conditions, the use of texture generated by ControlNet~\cite{zhang2023adding} often results in inconsistencies with the original mesh geometry, introducing ambiguity during training.
Alternatively, if the untextured meshes are rendered directly without texture synthesis, the resulting RGB images used as condition inputs tend to lose geometric information.
In our approach, we render high-quality RGB images for textured assets, and for untextured assets, we render corresponding normal maps. These RGB images or normal maps are then mixed as conditioning inputs for training the image-to-3D generation model, eliminating inconsistencies during training by providing accurate and geometry-aligned supervision. Moreover, the use of normal maps helps reduce the influence of high-frequency information typically present in texture data, enabling the model to learn more fidelity and structurally accurate 3D shapes.
\subsection{Linear Attention}\label{sec:Linear Attention}
\begin{figure}
    \centering
    \includegraphics[width=1\linewidth]{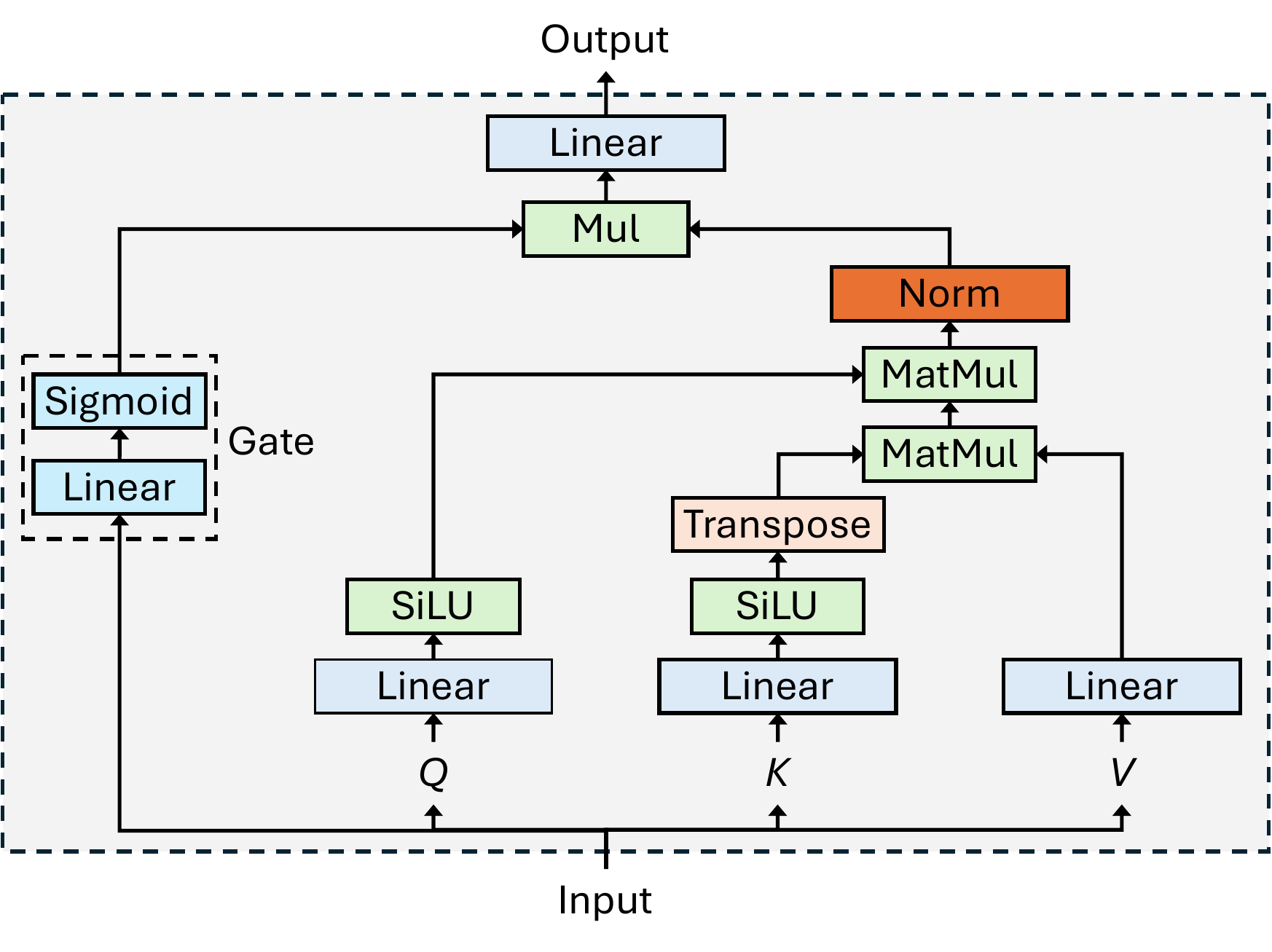}
    \vspace{-2em}
    \caption{The linear attention block used in our flow model.}
    \label{fig:linear_attention}
\end{figure}
\begin{table}[t]
\centering
\caption{Quantitative comparison of VAE reconstruction performance. CD. indicates Chamfer Distance, F.S. indicates F-Score, and N.C. indicates Normal Consistency.}
\resizebox{0.95\linewidth}{!}{
\begin{tabular}{lllllr}
\toprule
Methods & Token Num.& Strategy &CD.($\downarrow$) & F.S.($\uparrow$) & N.C.($\uparrow$) \\
\midrule
TripoSG &2,048&Trained& 4.51 & 0.999 & 0.958  \\
Ours & 4,096&Trained& 4.32& 0.999& 0.963\\
Ours & 16,384&Trained&4.15& 0.999& 0.970\\
Ours & 32,768 & Expanded&4.04&0.999& 0.973\\
Ours & 65,536&Expanded&3.98& 0.999& 0.974\\
\bottomrule
\end{tabular}
}
\label{table:vae_recon}
\end{table}
Existing approaches commonly follow the DiT architecture, employing softmax self-attention within each transformer block. However, when the number of tokens significantly exceeds the dimensionality of each token, linear attention becomes a more efficient alternative.
In the text-to-image generation field, SANA~\cite{xie2024sana} has taken the lead in incorporating linear attention into the DiT framework to accelerate high-resolution generation. However, image tokens exhibit strong local spatial correlations. A naive application of linear attention, while computationally efficient, leads to a loss of these local features. To mitigate this, SANA replaces the original feed-forward network (FFN) with a convolutional MixFFN to recover locality, at the cost of offsetting the speed gains brought by linear attention.
In contrast, following the design principles of exist works~\cite{zhang20233dshape2vecset,li2025triposg}, our approach compresses raw 3D data into 1D token sequences. Each token inherently encodes global structural information, thus avoiding the locality degradation issues encountered in 2D image representations.
Building on this insight, we incorporate linear attention following LightNet~\cite{qin2024you}, into our image-to-3D generation model.
As shown in Fig.\ref{fig:linear_attention}, after the linear projection layers, the key (K) is activated by a SiLU function and then multiplied with the value (V) via a matrix multiplication. The result is subsequently multiplied with the query (Q), which has also been passed through a SiLU activation.
In contrast to the conventional linear attention used in SANA, we adopt a gating mechanism introduced by LightNet, which contributes to more stable performance.
This enables efficient training and inference with large token numbers, significantly accelerating the model without compromising the fidelity of the generated 3D shapes.
\subsection{Inference-Time Scaling}\label{sec:Inference Scaling}
Following the approach in ~\cite{ma2025inference}, we explore inference-time scaling by optimizing noise selection during inference sampling. This leads to improved generation quality with only modest computational overhead.
We adopt the zero-order strategy proposed in ~\cite{ma2025inference}: at each step, candidate noises are sampled near a pivot noise, and the best candidate (selected by a verifier) propagates forward. 
In our setting, the verification process involves comparing the geometric structure of the generated 3D samples with the input image. Unlike comparing texts or text-image pairs, verifying consistency between a single image and a generated 3D asset is inherently more challenging, due to the modality gap and the lack of a general-purpose verifier. This raises difficulties in directly applying inference-time scaling to image-to-3D tasks.
To address this, we render front-view normal maps of the generated 3D assets and compare them with pseudo-reference normal maps derived from the input image to validate generation quality, as normal maps are highly correlated with the perceptual quality of 3D assets. Specifically, we use an off-the-shelf image-to-normal estimator ~\cite{ye2024stablenormal} to derive these reference maps. For verification, we compute the cosine similarity between DINOv2 ~\cite{oquab2023dinov2} feature embeddings extracted from the rendered and reference normal maps. This reward signal correlates well with human perception and captures fine-grained geometric details.

\section{Experiments}
We follow the data selection and processing strategies used in TripoSG~\cite{li2025triposg}, leveraging the Objaverse(-XL) dataset to curate our training data. This filtering process yields approximately 2 million high-quality samples for model training.
For VAE training, our implementation is primarily based on the open-source design of TripoSG, with enhancements to the representation and token scaling strategy.
The flow model training draws inspiration from ~\cite{li2025triposg,qin2024you,xiang2024structured}, while the core modules proposed in this paper are implemented independently as part of our pipeline. 
Due to the page limit, more training details are provided in the supplementary materials.
\subsection{VAE Reconstruction Performance}
\begin{figure}[t]
    \centering
    \includegraphics[width=1\linewidth]{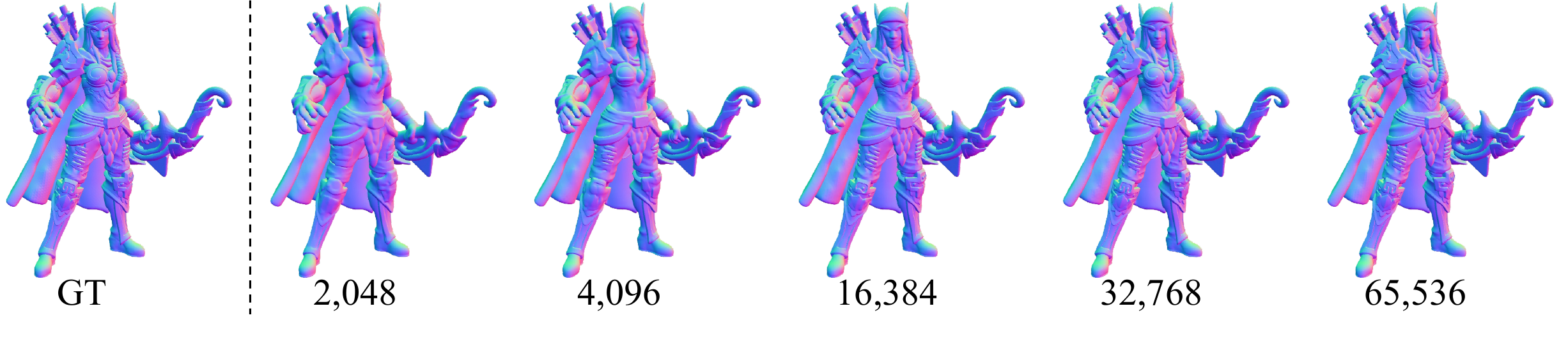}
    \vspace{-2em}
    \caption{Visualization of reconstruction results under different token numbers.}
    \label{fig:vae_recon}
\end{figure}

\begin{table}[t]
\centering
\caption{Comparison of reconstruction performance with other VAE methods, CD. indicates Chamfer Distance, F.S. indicates F-Score, and N.C. indicates Normal Consistency.}
\resizebox{1\linewidth}{!}{
\begin{tabular}{lrrr}
\toprule
Methods &CD.($\downarrow$) & F.S.($\uparrow$) & N.C.($\uparrow$)\\
\midrule
Direct3D~\cite{wu2024direct3d} &	5.08&	0.998&	0.913\\
CraftsMan3D-DoraVAE~\cite{li2024craftsman3d} &	4.72&	0.999&	0.931\\
TripoSG~\cite{li2025triposg}&	4.68&	0.999&	0.936\\
Dora-VAE-1.1~\cite{chen2025dora}&	4.52&	0.999&	0.949\\
Hunyuan3D-2.1-Shape~\cite{hunyuan3d2025hunyuan3d}&	4.37&	0.999&	0.960\\
Ours&	4.24&	0.999&	0.972\\
\bottomrule
\end{tabular}
}
\label{table:vae_recon_comparison_with_others}
\end{table}

Since the VAE training is implemented based on the open-source TripoSG~\cite{li2025triposg} framework, the reconstruction performance is primarily compared with TripoSG. As shown in Tab.~\ref{table:vae_recon}, we compare the reconstruction quality of our VAE with different token numbers to that of TripoSG. Among them, the results of 4,096 and 16,384 tokens are obtained through VAE model training, while the results for 32,768 and 65,536 tokens are expanded based on token scaling in the inference process.
From the results in the table, we can observe: 
Our model consistently achieves better reconstruction quality than TripoSG across different token numbers; 
The most significant improvement occurs when increasing the token number from 2,048 to 4,096 and 16,384;
Expanding to larger token numbers also leads to noticeable performance gains, indicating the model’s ability to generalize beyond the training token number setting.
Additionally, as illustrated in Fig.~\ref{fig:vae_recon}, we visualize the results of a complex case under different token numbers by rendering the normal maps of the reconstructed 3D models. It is clearly observable that as the number of tokens increases, the reconstruction quality improves progressively and approaches the ground truth. 
We also evaluated \MethodName{} and other methods that can obtain the VAE weights version on the Toys4K~\cite{stojanov2021using} dataset, and the results are as shown in Tab.~\ref{table:vae_recon_comparison_with_others}. Our VAE has a slightly leading performance due to the large number of input points and tokens.

\subsection{Image-to-3D Generation Performance.}
\begin{table}[t]
\centering
\caption{User study of image-to-3D generation.}
\resizebox{0.8\linewidth}{!}{
\begin{tabular}{lrr}
\toprule
Methods & Image-3D Alignment & Overall Quality \\
\midrule
Hunyuan3D-2&19.4&15.3 \\
TripoSG &21.2 &19.9\\
Hi3DGen &15.7 &29.2\\
Ours &43.7&35.6 \\
\bottomrule
\end{tabular}
}
\label{table:user_study}
\end{table}

\begin{table*}[h]
\centering
\caption{Comparison of image-to-3D generation performance with other methods, FD indicates Frechet Distance and KD indicates Kernel Distance.}
\vspace{-1em}
\resizebox{0.8\linewidth}{!}{
\setlength{\tabcolsep}{6pt}
\begin{tabular}{lrrrrrr}
\toprule
Methods & CLIP($\uparrow$) &$FD_{incep}(\downarrow)$&$KD_{incep}(\downarrow)$&$FD_{dinov2}(\downarrow)$&$KD_{dinov2}(\downarrow)$&$FD_{point}(\downarrow)$\\
\midrule
CraftsMan3D-1.5~\cite{li2024craftsman3d}&	82.11&	10.35&	0.11&	80.11&	0.98&	2.87\\
Trellis~\cite{xiang2024structured}&	85.89&	9.21&	0.03&	68.78&	0.77&	2.00\\
Step1X-3D~\cite{li2025step1x}&	85.19&	9.31&	0.03&	69.92&	0.76&	2.03\\
Hunyuan3D-2~\cite{zhao2025hunyuan3d}&	85.01&	8.99&	0.03&	69.90&	0.69&	1.89\\
TripoSG~\cite{li2025triposg}&	87.34&	8.02&	0.02&	64.04&	0.63&	1.77\\
Hi3DGen~\cite{ye2025hi3dgen}&	87.90&	7.46&	0.02&	62.85&	0.55&	1.54\\
Ours&	88.21&	7.29&	0.02&	62.34&	0.50&	1.38 \\
\bottomrule
\end{tabular}
}
\vspace{1em}
\label{table:generation_comparison_with_others}
\end{table*}

\begin{figure*}
    \centering
    \includegraphics[width=1.02\linewidth]{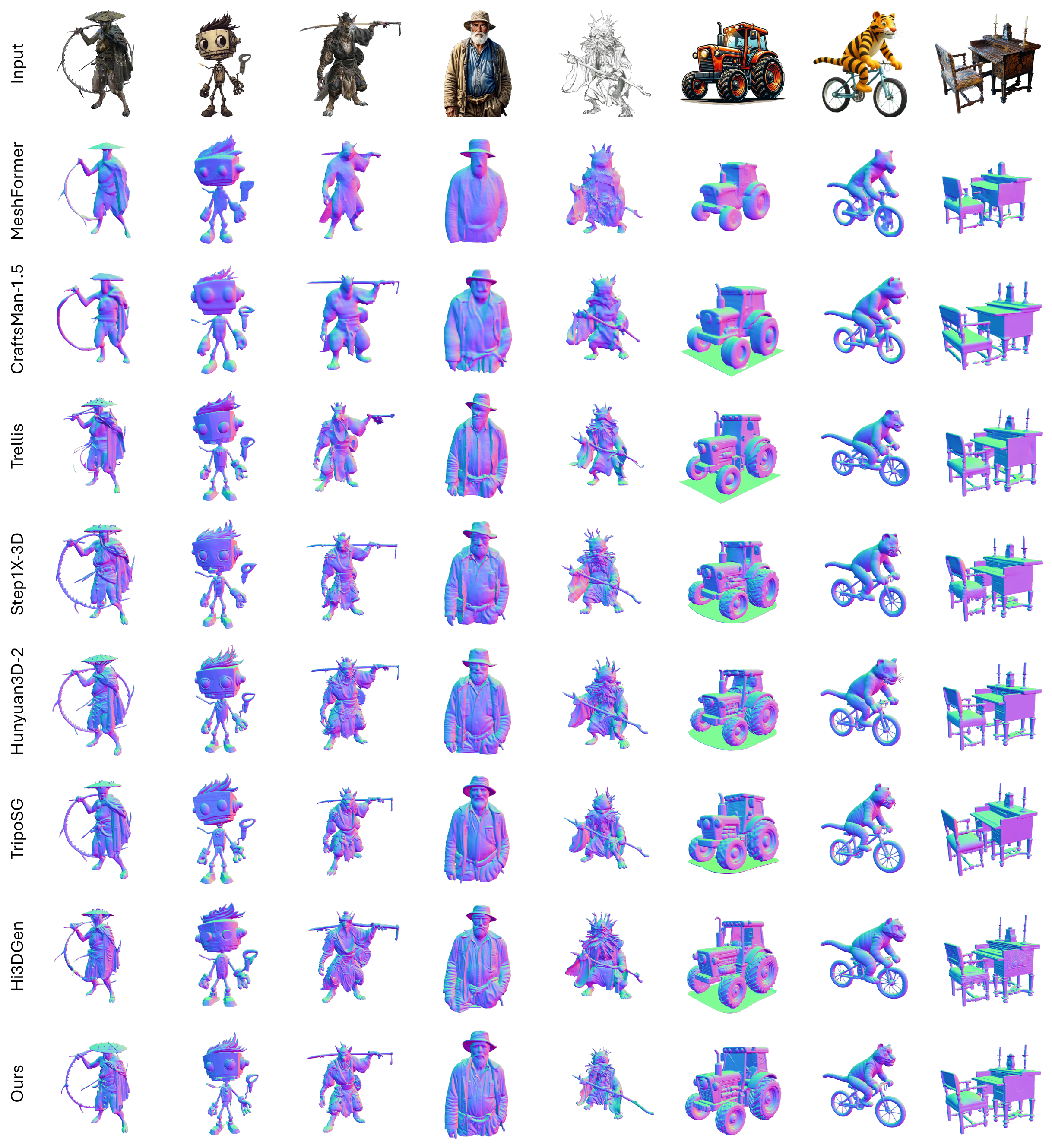}
    \caption{Qualitative comparison between our method and prior state-of-the-art approaches on image-to-3D generation results. We render normal maps from the generated 3D models for visualization to highlight the geometric details. We refer to TripoSG for these test cases, which are different from the commonly used character or toy images in 3D generation, to conduct complex details and diverse tests to better show the real capabilities of the model.}
    \label{fig:shapegen_comparison}
\end{figure*}

\begin{figure*}
    \centering
    \vspace{-0.5em}
    \includegraphics[width=1\linewidth]{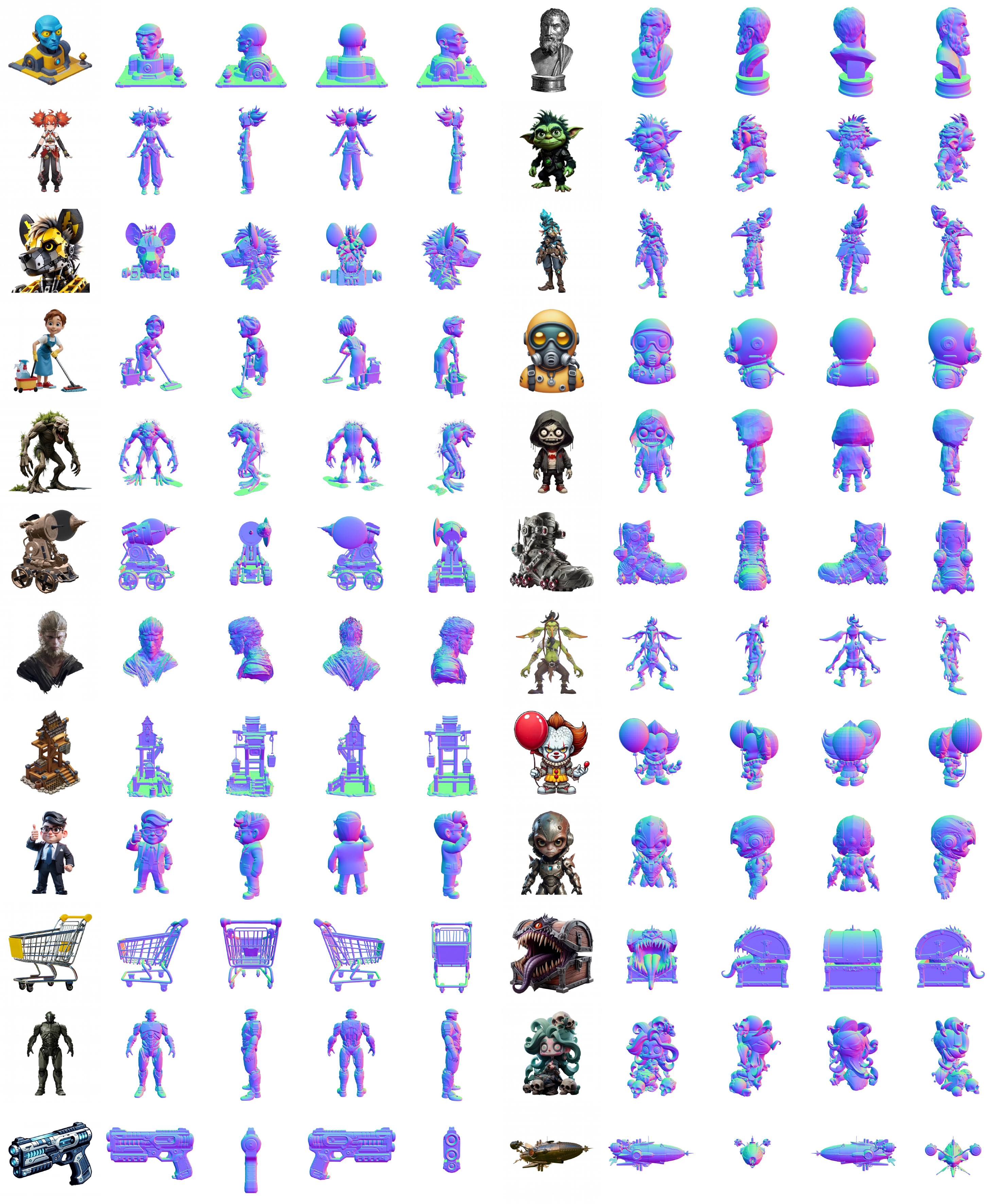}
    \vspace{-1.5em}
    \caption{More visualizations of image-to-3D generation results produced by our method.}
    \label{fig:shapegen_demo_show}
\end{figure*}

Due to limited training resources, our image-to-3D flow model is trained progressively, starting from 2,048 tokens and scaling up to 16,384 tokens. We compare the final model against existing open-source image-to-3D diffusion and flow-based methods. As shown in Fig.~\ref{fig:shapegen_comparison}, our approach consistently outperforms prior works~\cite{liu2024meshformer,li2024craftsman,xiang2024structured,li2025step1x,zhao2025hunyuan3d,li2025triposg,ye2025hi3dgen} in terms of alignment with the input image, structural detail fidelity, and overall visual quality.
For example, compared to Hi3DGen~\cite{ye2025hi3dgen}, our method achieves better consistency, which may be attributed to Hi3DGen’s reliance on normal maps as an intermediate representation, which struggles to preserve accurate details in complex cases. Compared to methods such as TripoSG~\cite{li2025triposg} and Hunyuan3D-2~\cite{zhao2025hunyuan3d}, our model demonstrates richer and more consistent details across the generated 3D assets.
To further evaluate the top-4 methods (Hunyuan3D-2, TripoSG, Hi3DGen, and ours), we conducted a user study. Specifically, we collected 125 input images and performed inference using all four methods. 10 human raters were then assess each output based on image consistency and overall quality, selecting the best result for each input. Final scores were averaged across raters. As shown in Tab.~\ref{table:user_study}, our approach outperforms all baselines in both image-3D alignment and overall quality metrics.
We also follow the Trellis~\cite{xiang2024structured} evaluation metric to evaluate the image-to-3D performance of the model weight available methods in Fig.~\ref{fig:shapegen_comparison} by randomly rendering an image on the Toys4k~\cite{stojanov2021using} dataset, and the results are shown in Tab.~\ref{table:generation_comparison_with_others}. From the perspective of quantitative indicators, our \MethodName{} also has advantages.
Fig.~\ref{fig:shapegen_demo_show} presents additional generation results from our model, covering a diverse range of object categories and styles. These results highlight the strong generalization ability and high-quality generation capabilities of our method.

\section{Ablation and Discussion}
\subsection{VAE Representation and Supervision}
\begin{figure}[t]
    \centering
    \includegraphics[width=1\linewidth]{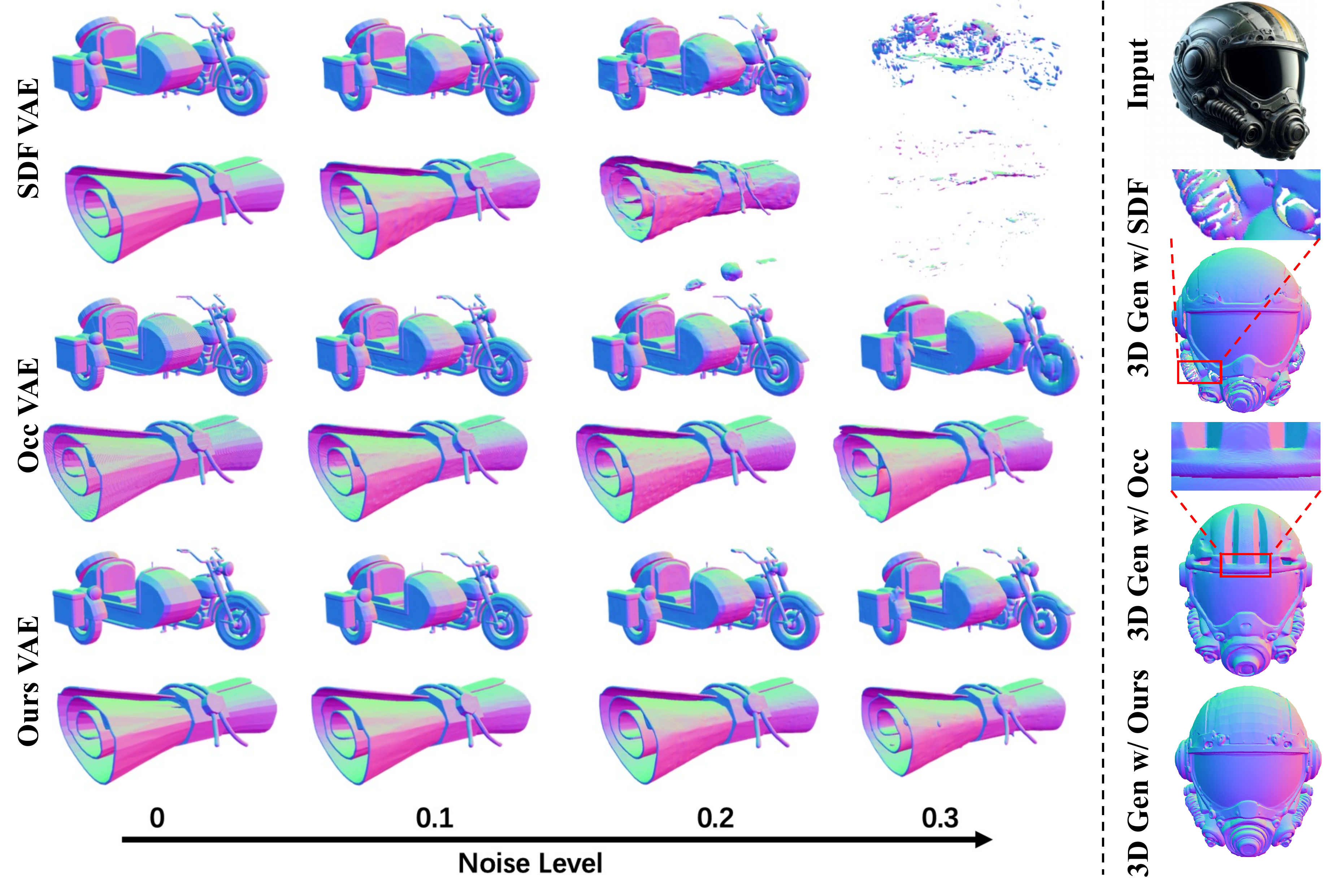}
    \vspace{-1em}
    \caption{Comparison of different 3D representations on VAE reconstruction and image-to-3D generation performance.}
    \vspace{-1em}
    \label{fig:supervison_ablation}
\end{figure}
Few works have explored how 3D VAE-compressed latents affect generative model learnability. Despite similar reconstruction quality across representations, their noise robustness varies, significantly impacting downstream generation performance.
We compare our VAE with BCE loss against two commonly used alternatives: occupancy with BCE loss (Occ VAE) and SDF with L1/L2 loss (SDF VAE). As shown in the left part of Fig.~\ref{fig:supervison_ablation}, we evaluate the reconstruction quality of VAEs trained with different representations and visualize the effects of injecting varying levels of noise into their latent features during decoding. The results reveal that occupancy representation suffer from pronounced aliasing or "staircasing" artifacts. While SDF produces smoother surfaces, its latent features are more sensitive to noise, even smaller perturbations can severely distort the decoded shape, making it less suitable for generative modeling.
Our VAE combines the advantages of both occupancy and SDF. It inherits the smooth surface properties of SDF and the noise robustness of occupancy, resulting in more stable latent spaces.
On the right side of Fig.~\ref{fig:supervison_ablation}, we further visualize image-to-3D results from generation models trained with these VAEs. Models trained with SDF tend to generate fragmented thin shell, while those trained with occupancy exhibit "staircasing" artifacts. In contrast, our VAE leads to superior overall quality and structural consistency, particularly in handling fine, thin geometric details.

\subsection{Normal vs. Image Condition}
\begin{figure}
    \centering
    \includegraphics[width=1\linewidth]
    {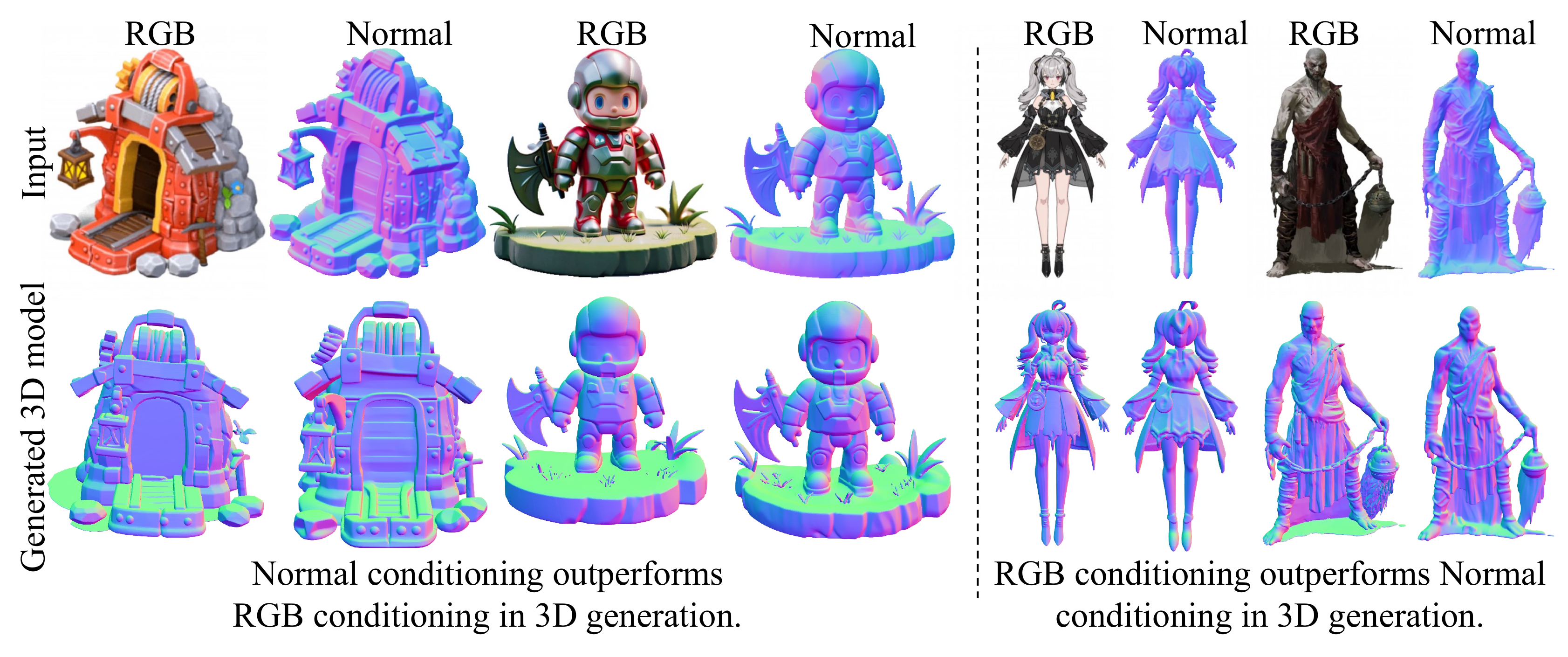}
    \vspace{-1em}
    \caption{Comparison of 3D asset generation results using normal maps vs. RGB as conditioning inputs.}
    \vspace{-1em}
    \label{fig:rgb_normal_comparison}
\end{figure}
During training, we render RGB images as conditioning inputs for textured 3D assets and normal maps for untextured ones, enabling our flow model to accept either RGB or normal maps as input for 3D generation. Intuitively, normal maps may offer better geometric detail generation, as they eliminate high-frequency texture and color information that could interfere with geometry learning.
However, in actual use, acquiring high-quality and accurate normal maps is challenging. When relying on RGB-to-normal prediction models, the estimated normal maps are often inaccurate for complex images, leading to degraded 3D generation quality when used as conditions.
As shown in Fig.~\ref{fig:rgb_normal_comparison}, we compare cases where normal map conditioning performs better (left) and worse (right) than RGB conditioning. The normal maps are predicted from the corresponding RGB inputs using StableNormal~\cite{ye2024stablenormal}. When the predicted normal map is accurate, it leads to better 3D generation than RGB conditioning. Conversely, inaccurate normal maps result in poorer outcomes.
Thus, the effectiveness of using normal maps as conditioning inputs depends heavily on the accuracy of the predicted normals.

\subsection{Linear Attention}
\begin{figure}[t]
    \centering
    \includegraphics[width=1\linewidth]{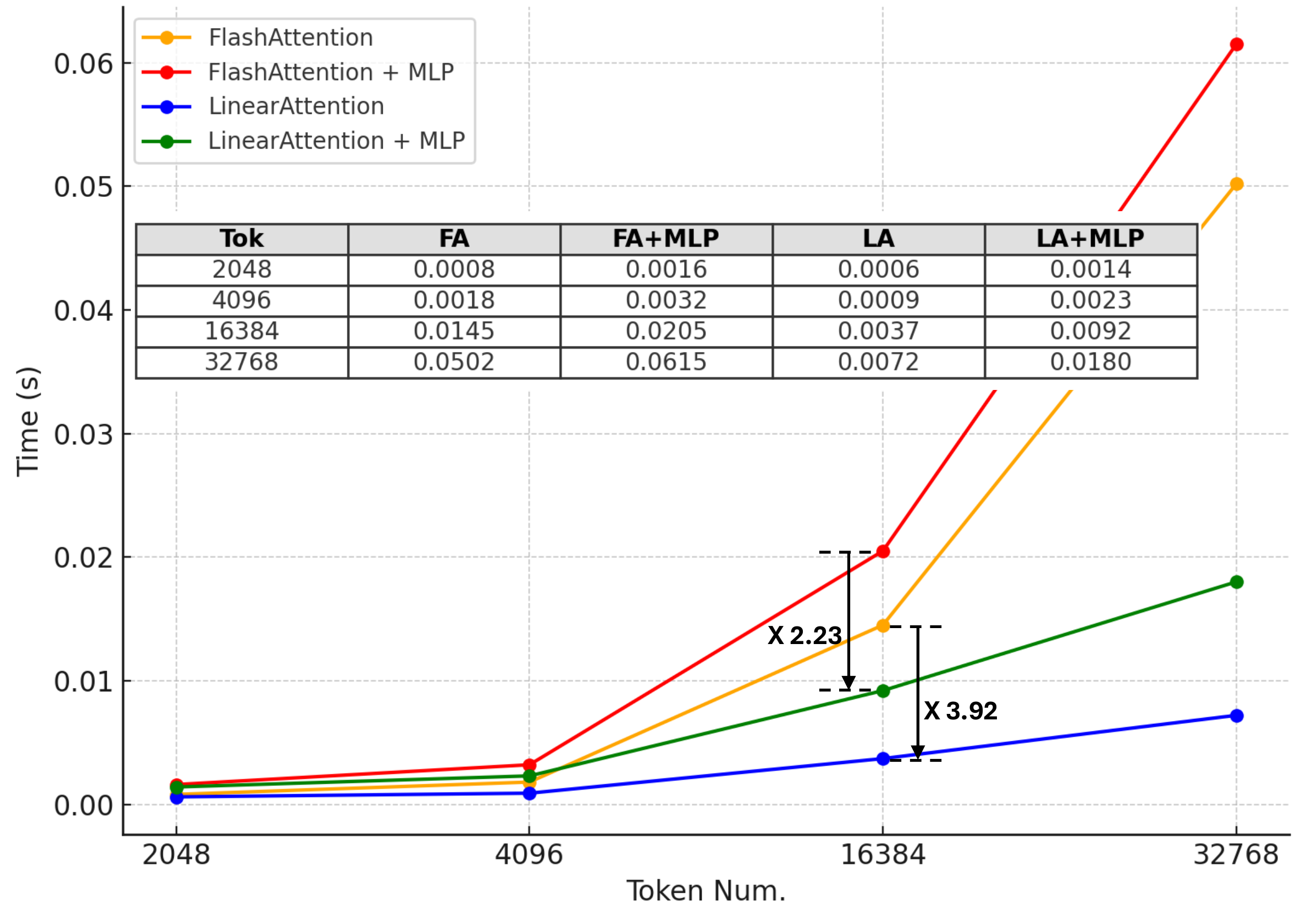}
    \vspace{-1em}
    \caption{Training speed of a single linear attention operation under different token numbers (in seconds). `FA' and `LA' denote Flash and Linear Attention.}
    \label{fig:linear_attention_performance}
\end{figure}
In the domain of 3D asset generation, our work is the first to introduce linear attention operations into a generative models. As shown in Fig.~\ref{fig:linear_attention_performance}, the performance benefits of linear attention become increasingly prominent as the token number grows. In particular, when the number of tokens reaches 32,768, far exceeding the token dimension of 64, pure linear attention achieves a $6.97\times$ speedup over Flash Attention.
A common consideration is that linear attention block includes an FFN component composed of MLP layers, whose computation cost grows with the number of tokens, potentially reducing efficiency. However, when token number far exceeds token dimension, linear attention still offers notable speed advantages. Specifically, when the token number is 16,384, the linear attention block achieves a $2.23\times$ speedup over the Flash Attention block; when the token count is 32,768, the speedup increases to $3.42\times$.
In our final training phase, the flow model is trained with 16,384 tokens, making the speed advantage of linear attention impactful. It is also important to mention that replacing the cross-attention used for conditioning with linear attention leads to a slight performance drop. Therefore, all results presented in the paper are based on models that use linear attention in self-attention blocks only, not in cross-attention.

\subsection{Inference-Time Scaling}
\begin{figure}[t]
    \centering
    \includegraphics[width=1\linewidth]{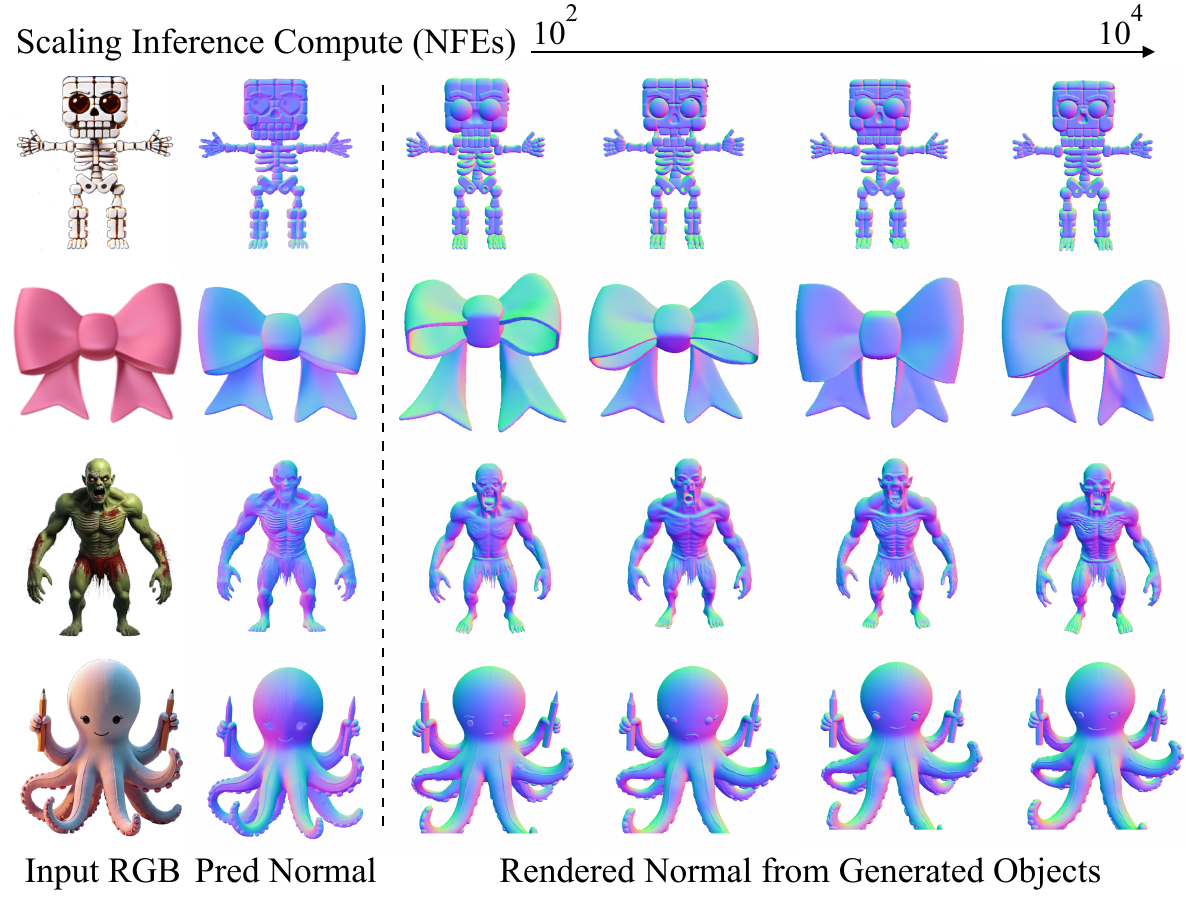}
    \vspace{-1em}
    \caption{Visualization of inference-time scaling: left-input RGB images and the corresponding pseudo normal maps; right-front-view normal renderings from the generated 3D assets under increasing inference budgets.}
    \label{fig:inference_scaling_examples}
\end{figure}

We demonstrate inference-time scaling using single RGB images as input. The base denoising process is configured with a 50-step (NFE) diffusion schedule. For search-based inference scaling, we set a total budget of 12,800 NFEs, corresponding to 32 rounds of zero-order search with 8 candidate samples per round.
As illustrated in Fig.~\ref{fig:inference_scaling_examples}, increasing the inference budget leads to improved geometric alignment and finer details in the generated results—such as the more accurate bow shape in the second row and enhanced facial expression of the octopus in the last row. The quantitative results provided in supplementary also highlights the promise of applying inference scaling to the image-to-3D generation task.

\section{Conclusion}

We introduced \MethodName{}, a new method for generating high-quality 3D shapes from a single image. By improving supervision strategies, scaling up data resolution and token number, using RGB or Normal maps for conditioning, adopting linear attention, and applying an inference-time scaling strategy, \MethodName{} addresses key issues like missing details and surface artifacts in existing methods.
Each component improves quality, and together they enable \MethodName{} to achieve state-of-the-art image-to-3D generation, with practical applicability to real-world 3D tasks.
Notably, \MethodName{} has yet to incorporate Dora’s~\cite{chen2024dora} edge-enhanced token sampling, which could further boost detail and quality. And with only 1.5B parameters is far smaller than typical 10B+ models of image and video generation. Future work will explore scaling model parameters, edge-aware sampling, and larger datasets to further improve image-to-3D generation.

\begin{acks}
This work was supported by the JC STEM Lab of AI for Science and Engineering, funded by The Hong Kong Jockey Club Charities Trust, the Research Grants Council of Hong Kong (Project No. CUHK14213224).
\end{acks}

\bibliographystyle{ACM-Reference-Format}
\bibliography{sample-bibliography}

\end{document}